\documentclass[sigconf]{acmart}
\AtBeginDocument{%
  \providecommand\BibTeX{{%
    \normalfont B\kern-0.5em{\scshape i\kern-0.25em b}\kern-0.8em\TeX}}}

\copyrightyear{2021} 
\acmYear{2021} 
\setcopyright{acmlicensed}\acmConference[MM '21]{Proceedings of the 29th ACM International Conference on Multimedia}{October 20--24, 2021}{Virtual Event, China}
\acmBooktitle{Proceedings of the 29th ACM International Conference on Multimedia (MM '21), October 20--24, 2021, Virtual Event, China}
\acmPrice{15.00}
\acmDOI{10.1145/3474085.3475692}
\acmISBN{978-1-4503-8651-7/21/10}

\definecolor{ForestGreen}{rgb}{0.0, 0.5, 0.0}
\definecolor{brandeisblue}{rgb}{0.0, 0.44, 1.0}
\definecolor{cadmiumred}{rgb}{0.89, 0.0, 0.13}
\usepackage[ruled,vlined]{algorithm2e}
\begin{document}

%%
%% The "title" command has an optional parameter,
%% allowing the author to define a "short title" to be used in page headers.
\title{Exploiting BERT For Multimodal Target Sentiment Classification Through Input Space Translation}

%%
%% The "author" command and its associated commands are used to define
%% the authors and their affiliations.
%% Of note is the shared affiliation of the first two authors, and the
%% "authornote" and "authornotemark" commands
%% used to denote shared contribution to the research.
\author{Zaid Khan, Yun Fu}
\orcid{0000-0003-0743-2992}
\email{{khan.za, y.fu}@northeastern.edu}
\affiliation{%
  \institution{Northeastern University}
  \city{Boston}
  \state{Massachusetts}
  \country{USA}
}

%%
%% By default, the full list of authors will be used in the page
%% headers. Often, this list is too long, and will overlap
%% other information printed in the page headers. This command allows
%% the author to define a more concise list
%% of authors' names for this purpose.
% \renewcommand{\shortauthors}{Trovato and Tobin, et al.}

%%
%% The abstract is a short summary of the work to be presented in the
%% article.
\begin{abstract}
Multimodal target/aspect sentiment classification combines multimodal sentiment analysis and aspect/target sentiment classification. The goal of the task is to combine vision and language to understand the sentiment towards a target entity in a sentence. Twitter is an ideal setting for the task because it is inherently multimodal, highly emotional, and affects real world events. However, multimodal tweets are short and accompanied by complex, possibly irrelevant images. We introduce a two-stream model that translates images in input space using an object-aware transformer followed by a single-pass non-autoregressive text generation approach. We then leverage the translation to construct an auxiliary sentence that provides multimodal information to a language model. Our approach increases the amount of text available to the language model and distills the object-level information in complex images. We achieve state-of-the-art performance on two multimodal Twitter datasets without modifying the internals of the language model to accept multimodal data, demonstrating the effectiveness of our translation. In addition, we explain a failure mode of a popular approach for aspect sentiment analysis when applied to tweets. Our code is available at \textcolor{blue}{\url{https://github.com/codezakh/exploiting-BERT-thru-translation}}.
% Social media is an ideal setting for the multimodal sentiment analysis task because of its centrality and importance to real world events and inherent multimodality. 
% In contrast to classical multimodal sentiment analysis, which has decades of fundamental science research in human facial/auditory emotional expression to draw insights from, there is little fundamental research on multimodal sentiment expression in the social media setting. 
% We fill this literature gap by analyzing and describing patterns in the use of images in tweets to convey emotion. Motivated by the mined patterns, we hypothesize that high level object/scene context in images is highly relevant for capturing emotion, in contrast to prior methods, which use visual features from pretrained image classification models. 
% We demonstrate the effectiveness of a surprisingly simple method for providing rich image context to large language models without making architectural modifications. 
% We show that a popular data augmentation for sentiment analysis fails in the short-text social media setting, and develop an alternative that is 3x faster. 
% We demonstrate the wide applicability of our methods across four multimodal tweet datasets, three tasks, and multiple backbones. 
% Our study of multimodal sentiment expression in tweets provides insight for future research, while the simplicity and strong empirical results of our approach will be attractive to practitioners.
\end{abstract}

%%
%% The code below is generated by the tool at http://dl.acm.org/ccs.cfm.
%% Please copy and paste the code instead of the example below.
%%


%%
%% Keywords. The author(s) should pick words that accurately describe
%% the work being presented. Separate the keywords with commas.

%% A "teaser" image appears between the author and affiliation
%% information and the body of the document, and typically spans the
%% page.
\begin{teaserfigure}
  \includegraphics[width=\textwidth]{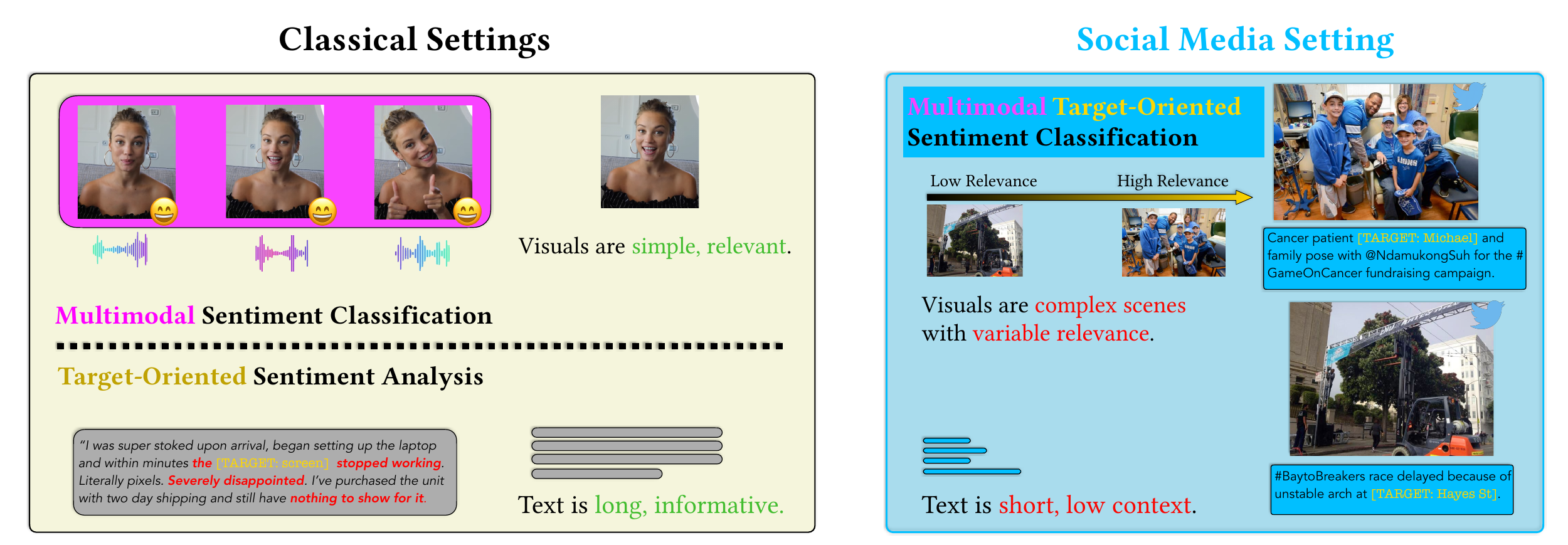}
  \caption{The challenges of multimodal target/aspect sentiment classification on social media compared to classical settings.}
  \label{fig:teaser}
\end{teaserfigure}

%%
%% This command processes the author and affiliation and title
%% information and builds the first part of the formatted document.
\maketitle

\section{Introduction}
\begin{figure*}
    \centering
    \includegraphics[width=\textwidth]{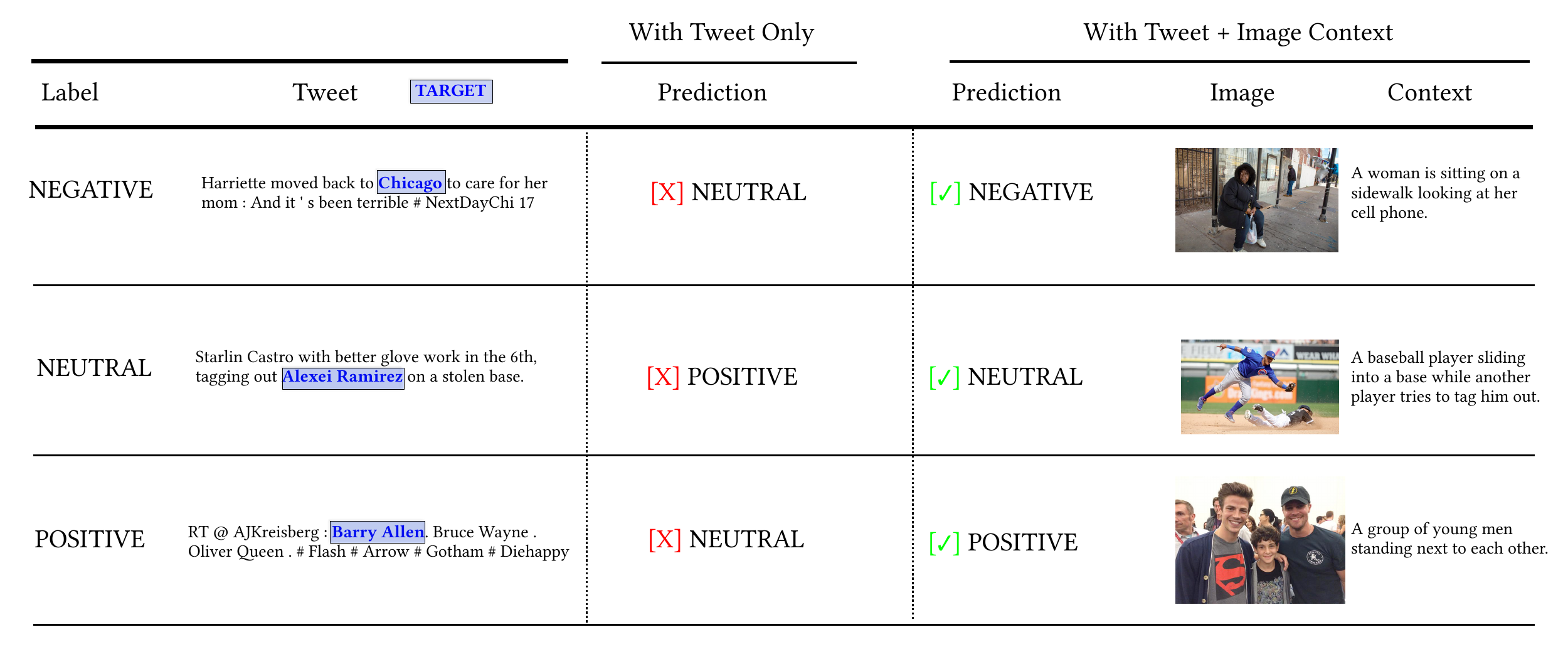}
    \caption{Examples the text-only BERT predicts incorrectly, but our proposed method gets correct. In Row 2, notice how the proposed method uses the image context to recognize the target is the player being tagged out and downgrades the sentiment.}
    \label{fig:qualitiative-proof}
\end{figure*}

The spread of emotional narratives is conjectured to drive large-scale events \cite{shillerNarrativeEconomics2017} and Twitter is a critical venue for spreading emotional narratives \cite{jacksonHashtagActivismNetworksRace2020} and misinformation \cite{kusenPoliticsSentimentsMisinformation2018, vosoughiSpreadTrueFalse2018}. Understanding the spread of sentiment on the scale of social media requires computational approaches to emotion understanding. A holistic computational understanding of emotion requires comprehension of \textit{aspect/target} - towards \textit{what} is the sentiment directed? In addition, the inherent multimodality of tweets requires models capable of multimodal sentiment understanding, specifically vision/language, to leverage the emotional information and additional context carried by the images accompanying tweets.

While aspect sentiment analysis and multimodal sentiment analysis are established fields, multimodal aspect sentiment analysis - a combination of the two - is relatively new. Directly applying methods developed for aspect sentiment analysis or multimodal sentiment analysis to multimodal tweets is challenging for the following reasons: (1) tweets are shorter and less informative than the review texts commonly used for aspect sentiment analysis, (2) the visual modality is much more likely to be irrelevant/noisy for tweets than for videos of human speakers commonly used for multimodal sentiment analysis, (3) the visual scenes accompanying tweets are significantly more complex than videos of single human speakers. 

We propose a model to address these challenges. Our approach adapts a transformer architecture for object detection to generate text instead, then uses the resulting model to translate images into the input space of a pretrained language model. We then feed the translated image into a BERT language model by constructing an auxillary sentence, and use the resulting encoding for multimodal aspect sentiment analysis. This solves two problems: first, we increase the amount of text available for the language model, and second, the translation preserves complex scene information appropriate for understanding social media images. Finally, the translation renders the fusion relatively interpretable, because the image is expressed in natural language. 

Our work is similar in spirit to \cite{liVisualQuestionAnswering2019}, who proposed using image captions for the visual question answering task. However, \cite{liVisualQuestionAnswering2019} requires external information, does not leverage recent large language models, and is for the VQA task rather than sentiment classification. We draw inspiration from BERT-Pair-QA \cite{sunUtilizingBERTAspectBased2019}, which constructs an auxillary sentence to aid aspect sentiment classification. Unlike their auxillary sentence, we use the auxillary sentence to pass multimodal information into BERT. In principle, our work has similarities to MCTN \cite{phamFoundTranslationLearning2019a}, which cyclically translates modalities for fusion, but we perform translation in the \textit{input space} rather than the feature space, and do not need a GAN. Our approach shares characteristics with TomBERT \cite{yuAdaptingBERTTargetOriented2019}, but unlike TomBERT, we perform multimodal fusion without modifying BERT, allowing us to achieve higher performance. We rely heavily on the architecture introduced by DETR \cite{carionEndtoEndObjectDetection2020} for end-to-end object detection with transformers, but we use the DETR layers to implement a non-autoregressive text generator rather than an object detector.
Our contributions are the following:
\begin{enumerate}
    \item A novel architecture for multimodal aspect sentiment analysis that translates in input space and utilizes an auxillary sentence for fusion through a large pretrained language model. 
    \item We investigate and explain the performance degradation of state-of-the-art aspect sentiment analysis models in the short-text twitter setting.
    \item We adapt the DETR \cite{carionEndtoEndObjectDetection2020} architecture for non-autoregressive text generation.
\end{enumerate}

\section{Related Work}

\subsection{Multimodal Sentiment Classification}
The goal of multimodal sentiment analysis is to regress or classify the overall sentiment of an utterance using acoustic, visual, and language cues. Because multimodal sentiment analysis is a large and well-established field, we direct the reader to \cite{baltrusaitisMultimodalMachineLearning2019, mogadalaTrendsIntegrationVision2020, poriaTipIcebergCurrent2020} for an overview of the field, and MISA \cite{hazarikaMISAModalityInvariantSpecific2020}, MAG \cite{rahmanIntegratingMultimodalInformation2020a}, and M3ER \cite{mittalM3ERMultiplicativeMultimodal2019} as representative of recent state of the art works. We restrict our scope to describing differences and similarities between our setting and the classical multimodal sentiment analysis setting. The primary difference is that multimodal sentiment analysis datsets such as \cite{zadehMOSIMultimodalCorpusa, bagherzadehMultimodalLanguageAnalysis2018b, parkComputationalAnalysisPersuasiveness2014, bussoIEMOCAPInteractiveEmotional2008a} traditionally focus on videos of a speaker with their face in focus. In this scenario, it is overwhelmingly likely the visual modality is relevant, in contrast to the social media setting, where images accompanying tweets can be completely irrelevant \cite{vempalaCategorizingInferringRelationship} or lack emotional content. In addition, multimodal sentiment analysis models operate on \textit{sequences} of multimodal data, whereas our setting is more similar to the vision-language setting, where only the language input is a sequence, and the accompanying modality is a single image. 

\subsection{Target/Aspect Sentiment Classification Without Architectural Modifications}
The goal of target/aspect sentiment classification is to classify the sentiment \textit{towards a target} mentioned in text. The power and widespread availability of large pretrained language models such as BERT\cite{luViLBERTPretrainingTaskAgnostic2019}, roBERTa \cite{liuRoBERTaRobustlyOptimized2019}, GPT-2 \cite{radfordLanguageModelsAre}, and XLNet\cite{yangXLNetGeneralizedAutoregressive2020} has resulted in pretrained language models dominating the field. Although architectural modifications to large pretrained language models have been successful \cite{zengLCFLocalContext2019}, studies have shown that architectural modifications to large language models are \cite{narangTransformerModificationsTransfer2021} brittle and often do not transfer across implementations and applications. In addition, \cite{mukherjeeReproducibilityReplicabilityAssessing2021a} finds that reproducing the results of modifications to BERT architectures is difficult.
However, the remarkable ability of transformers to adapt without architectural modifications \cite{luPretrainedTransformersUniversal2021}, as well as the practical desirability of deploying \cite{paleyesChallengesDeployingMachine2021} popular baseline architectures without modifications, has motivated research into exploiting large language models for target/aspect sentiment classification without architectural modifications. 

These methods either rely on data augmentation, novel inference procedures, enhanced training procedures \cite{xuBERTPostTrainingReview}, or a combination of the three. Our work fits most closely into this mold. The most related work to ours in the BERT-Pair model introduced by Sun \cite{sunUtilizingBERTAspectBased2019}. The BERT-Pair model converts the multi-class aspect sentiment classification problem into a series of binary classification problems by using BERT in sentence-pair mode. To form the second sentence of the pair, they construct an auxillary sentence that encodes label information as a query. \textbf{In contrast to BERT-Pair \cite{sunUtilizingBERTAspectBased2019}, we encode multimodal image context through translation into the auxillary sentence.} Furthermore, we do not require 3 passes for every training instance during training and inference, as BERT-Pair \cite{sunUtilizingBERTAspectBased2019} does. In addition, BERT-Pair \cite{sunUtilizingBERTAspectBased2019} cannot use multimodal data. 

\subsection{Transformers for Vision-Language Tasks}
Vision-language transformers are natively multimodal, taking in both language and visual input. The most similar to our work are two-stream encoders, such as ViLBERT \cite{luViLBERTPretrainingTaskAgnostic2019}. These use separate input streams for vision and language before fusing them at a later step. An ideal task for vision-language transformers is object detection and image captioning. Our caption transformer is directly inspired by DETR \cite{carionEndtoEndObjectDetection2020}, but \textbf{in contrast to DETR \cite{carionEndtoEndObjectDetection2020}, our DETR-based caption transformer predicts a natural language image description instead of a set of object detections}.
In this regard, our work makes use of a non-autoregressive text-generation approach - predicting all tokens in one pass - which makes it similar to works such as LaBERT \cite{dengLengthControllableImageCaptioning2020a} and Transform and Tell \cite{tranTransformTellEntityAware2020}, although we do not use a language model during the image captioning.

\subsection{Multimodal Target Oriented Sentiment Classification}
Multimodal target-oriented sentiment classification is a new field that combines multimodal sentiment analysis and target-oriented sentiment classification. The most common setting for this task is multimedia collected from online social media. Xu \cite{xuMultiInteractiveMemoryNetwork2019} introduced a Multi-Interactive Memory Network for multimodal aspect sentiment classification, and a private Chinese-language dataset called Multi-ZOL curated from review sites. The YouTubean dataset was introduced by Marrese-Taylor \cite{marrese-taylorMiningFinegrainedOpinions2017}, and contains target annotations 500 YouTube review videos.

The work most closely related to ours is TomBERT \cite{yuAdaptingBERTTargetOriented2019}, which introduces two multimodal tweet datasets with target annotations that we use, as well as a target-oriented multimodal BERT model that we compare to. TomBERT builds on top of the baseline BERT architecture by adding target-sensitive visual attention and adding more self-attention layers to capture inter-modality dynamics. \textbf{In contrast to TomBERT, we do not modify the baseline BERT model (other than one linear layer), and we do not add additional self-attention layers}.

\begin{table}
\caption{Key Differences Between The Proposed Model and Existing Approaches}
\label{tab:key-differences}
\begin{tabular}{@{}lccc@{}}
\toprule
                               & Proposed & TomBERT & VilBERT \\ \midrule
Doesn't Modify Language Model         & \textcolor{ForestGreen}{\checkmark}           & \textcolor{red}{X}       & \textcolor{red}{X}       \\
Trains End-to-End & \textcolor{red}{X} &  \textcolor{ForestGreen}{\checkmark}  &  \textcolor{ForestGreen}{\checkmark}  \\
Translates In Input Space      & \textcolor{ForestGreen}{\checkmark}         & \textcolor{red}{X}       & \textcolor{red}{X}      \\
Uses Object Level Information      & \textcolor{ForestGreen}{\checkmark}         & \textcolor{red}{X}       & \textcolor{ForestGreen}{\checkmark}      \\

Explictly Interpretable Fusion & \textcolor{ForestGreen}{\checkmark}         & \textcolor{red}{X}       & \textcolor{red}{X}       \\
\bottomrule
\end{tabular}
\end{table}

\section{Fusion and Sentiment Analysis Through Input Translation}

\subsection{Problem Definition}
Let $\mathcal{M}$ be a set of multimodal samples. Each sample $m_i \in \mathcal{M}$ consists of a sentence $S_i=(w_1, w_2, \ldots w_n)$, where $n$ is the number of words. Alongside the sentence is an image $I_i$, and an opinion target $T_i$, which is a subsequence of $S_i$. The opinion target is assigned a label $y_i \in \{\texttt{negative},\texttt{neutral},\texttt{positive} \}$. Our goal is to learn a function $f: (T_i, S_i, I_i) \rightarrow y_i$. In short, given a tweet like "Einstein is my favorite scientist, but I hate physics", and the posted image, the model should be able to predict \texttt{positive} for the target "Einstein" and \texttt{negative} for the target "physics".

\subsection{Overview}
\begin{figure*}
    \centering
    \includegraphics[width=\textwidth]{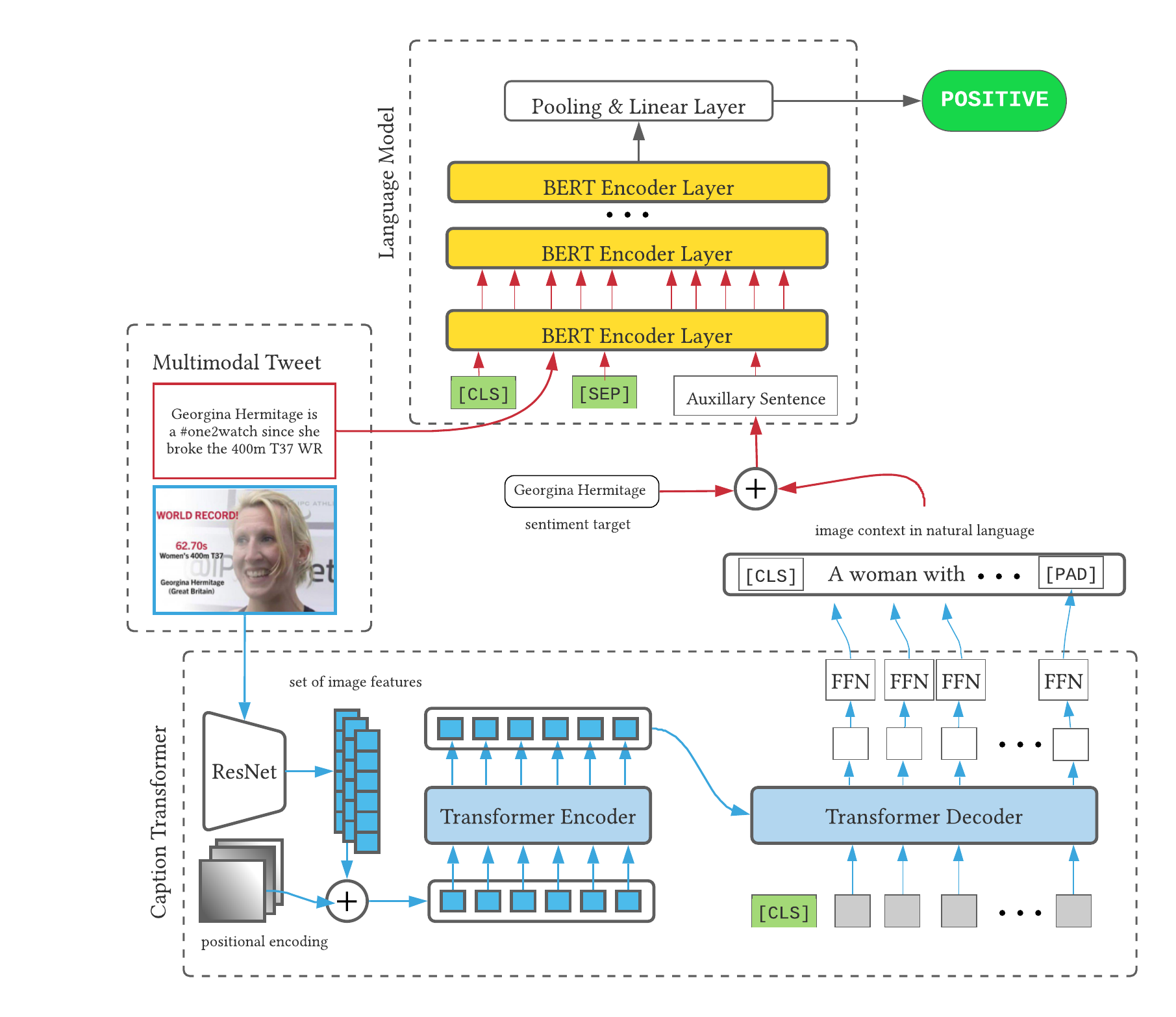}
    \caption{The architecture of our EF-CaTrBERT. The image input stream is in \textcolor{brandeisblue}{blue}, while the language input stream is in \textcolor{cadmiumred}{red}.}
    \label{fig:main-arch-diag}
\end{figure*}

Our model (Fig \ref{fig:main-arch-diag}) consists of two parts. Given a multimodal input sample $m_i = (T_i, S_i, I_i)$ consisting of a target $T_i$, input sentence $S_i$, and image $I_i$, we first pass the image through a captioning transformer. Let $3, H, W$ represent the number of channels, width, and height of an image respectively. The captioning transformer turns the image $I_i \in \mathbb{R}^{3\times H \times w}$ into a vector $\hat{I}_i \in \mathbb{N}_0^l$, where $l$ is the maximum output length of the captioning transformer. Thus, the image is converted from a 3-D tensor representing continuous data into a 1D vector of integers representing symbolic data. The tuple $\hat{I}_{i}, S_i, T_i \in \mathbb{N}_0^l$ representing image context, the sentence, and the target now reside in the same space, which is the vector space of $l$-dimensional vectors with elements in $N_0$. By sharing tokenizers and the vocabulary mapping symbolic words to elements in $N_0$ between the captioning transformer and the large language model, the language model can understand $\hat{I}_i \in \mathbb{N}_0^l$, which effectively becomes a natural language description of the image. We use the natural language image description $\hat{I}_i$ by constructing an auxillary $Aux_i$ sentence from the target $T_i$ and the natural language image description $\hat{I}_i$, and using the large language model in sentence-pair classification mode, feeding the pair $(S_i, Aux_i)$ through the language model to get a sentiment prediction $y_i$ for the target. The overall training procedure is described in Algorithm \ref{algo:training}.
\begin{algorithm}
\SetAlgoLined
\SetKwInOut{Input}{input}\SetKwInOut{Output}{output}
\Input{MS-COCO Dataset and Twitter-15/17 Dataset}
\Output{$\theta_{BERT}$, $\theta_{ResNet}$, $\theta_{FFN}$, $\theta_{Linear}$, $\theta_{DETRLayer}$}

\For{all epochs}{
\For{all batches in MS-COCO}{
  Forward text-image pairs through DETR backbone\;
  Compute loss $\mathcal{L}_1$ by Eq. \ref{eqn:catr-ce}\;
  Update $\theta_{FFN}$, $\theta_{DETRLayer}$, $\theta_{ResNet}$ using $\Delta \mathcal{L}_1.$
}
}
Initialize CaptionTransformer($\theta_{FFN}$, $\theta_{DETRLayer}$, $\theta_{ResNet}$) \;
\For{all epochs}{
\For{all batches in Twitter-15/17}{
  Forward images through CaptionTransformer \;
  Tokenize resulting description and tweet content. \;
  Construct auxillary sentence from description+target. \;
  Construct pair-input sentence for BERT. \;
  Forward sentence through BERT. \;
  Forward BERT pooler output through linear layer. \;
  Compute loss $\mathcal{L}_2$ by Eq. \ref{eqn:bert-output}\;
  Update $\theta_{BERT}$, $\theta_{Linear}$ using $\Delta \mathcal{L}_2$.

}
}
 \caption{Training and inference with EF-CaTrBERT.}
 \label{algo:training}
\end{algorithm}

\subsection{Input Translation with Caption Transformers}
\label{par:catr}
We propose to translate an image $I_i \in \mathbb{R}^{3\times H \times W}$ into an element in the input space $\hat{I}_i \in \mathbb{N}_0^l$ representing symbolic natural language input for a large language model, specifically BERT \cite{devlinBERTPretrainingDeep2019a}. 
\subsubsection{\textbf{Transformer Encoder}}
Given the input image $I_) \in \mathbb{R}^{3 \times H_0 \times W_0}$, we apply a CNN backbone (ResNet 101 \cite{heDeepResidualLearning2015a}) to generate an activation map $f \in \mathbb{C \times H \times W}$, where $C=2048$, and $H=W=\frac{H_0}{32}$. We then apply a 1x1 convolution to the channel dimension $C$ to reduce it to $d << C$, where $d=256$. The new feature map $z_0 \in \mathbb{R}^{d \times H \times W}$ is then flattened along the spatial dimensions, resulting in $z_0 \in \mathbb{R}^{d\times HW}$. The feature maps are then passed into a stack of DETR \cite{carionEndtoEndObjectDetection2020} encoder layers after augmentation with a fixed positional encoding. The DETR encoder architecture is a standard multi-head key, query, value architecture. We refer the reader to \cite{carionEndtoEndObjectDetection2020} for details, and provide a brief overview here. 

A DETR encoder layer consists of multiple \textit{attention heads}, each of which is parameterized by a weight tensor $T^{\prime} \in \mathbb{R}^{3 \times d^{\prime} \times d}$. The attention head first computes the \textit{key}, \textit{query}, and \textit{value} embeddings: 

\begin{equation}
[Q ; K ; V]=\left[T_{1}^{\prime}\left(X_{\mathrm{q}}+P_{\mathrm{q}}\right) ; T_{2}^{\prime}\left(X_{\mathrm{kv}}+P_{\mathrm{kv}}\right) ; T_{3}^{\prime} X_{\mathrm{kv}}\right]
\end{equation}

where $P_{\mathrm{q}} \in \mathbb{R}^{d \times N_{\mathrm{q}}}$ and $P_{\mathrm{kv}} \in \mathbb{R}^{d \times N_{\mathrm{kv}}}$ are position embeddings for the key and query respectively. Because the transformer is permutation invariant, the positional embeddings are added to each transformer encoder layer to help the transformer learn dependencies. The weight tensor $T^\prime$ is the concatenation of $T_{1}^{\prime}, T_{2}^{\prime}, T_{3}^{\prime}$. The attention then computes the attention weights, which represents the importance of each element in the key-value sequence to each element of the query sequence. Let $i$ be an index of the query sequence and $j$ be the index of the key-value sequence, then the attention weights $\alpha_{i,j}$ are given by 
$$
\alpha_{i, j}=\frac{e^{\frac{1}{\sqrt{d^{\prime}}} Q_{i}^{T} K_{j}}}{Z_{i}} \text { where } Z_{i}=\sum_{j=1}^{N_{\mathrm{kv}}} e^{\frac{1}{\sqrt{d^{\prime}}} Q_{i}^{T} K_{j}} .
$$
The output of the attention head is the weighted sum of the value sequence $V$, using the attention as weights. The $i-th$ row is given by $\operatorname{attn}_{i}\left(X_{\mathrm{q}}, X_{\mathrm{kv}}, T^{\prime}\right)=\sum_{j=1}^{N_{\mathrm{kv}}} \alpha_{i, j} V_{j}$. We then combine the single attention heads of a single encoder layer into a multi-head attention by concatenating the outputs of single attention heads $1 \ldots M$
$$
\begin{array}{l}
X_{\mathrm{q}}^{\prime}=\left[\operatorname{attn}\left(X_{\mathrm{q}}, X_{\mathrm{kv}}, T_{1}\right) ; \ldots ; \operatorname{attn}\left(X_{\mathrm{q}}, X_{\mathrm{kv}}, T_{M}\right)\right] \\
\tilde{X}_{\mathrm{q}}=\operatorname{layernorm}\left(X_{\mathrm{q-}}+\operatorname{dropout}\left(L X_{\mathrm{q}}^{\prime}\right)\right)
\end{array}
$$
followed by a linear projection and a residual connection. This is the standard formulation of key, query, value attention by \cite{vaswaniAttentionAllYou2017}.
\subsubsection{\textbf{Transformer Decoder}}
The decoder transforms $l$ embeddings of size $d$, where $l$ is the maximum sequence length the transformer can attend to. In contrast to DETR \cite{carionEndtoEndObjectDetection2020}, which inputs \textit{object queries} and trains the transformer with a set-matching loss, we use the decoder for non-autoregressive text generation by predicting a description for the input image in one forward pass. We generate a description as follows. Let $v^* = \operatorname{concat}(\operatorname{TokenID}{\texttt{[CLS]}}, zeros(l - 1))$ be the concatenation of the BERT token ID for the \texttt{[CLS]} token with the $l-1$ dimensional zero vector. The resulting vector $v^*$ becomes a prompt for the decoder, indicating the start of a sentence. We add a standard positional encoding to $v^*$ from \cite{carionEndtoEndObjectDetection2020} and forward it through the encoder to get $l$ embeddings of size $d$. In the decoder, we apply a 3-layer feedforward network with ReLU activations \cite{nairRectifiedLinearUnits} to predict a word from BERT's vocabulary at each query position. The token \texttt{[PAD]} is used to represent the absence of words at the end of the sentence.
At each position in the prompt sequence $v^*$ (except for the first position, which contains a control code), we predict a $30522$-dimensional probability distribution over BERT's vocabulary:

\begin{equation}
\label{eqn:catr-output}
       p(t|v^*,i]) = \operatorname{SoftMax}\textcolor{red}{(}W_3 \times R(W_2 \times R\textcolor{ForestGreen}{(}W_1 \times D_{ENC}(v^*)_i]\textcolor{ForestGreen}{)})\textcolor{red}{)}
\end{equation}

where $D_{ENC}(v^*)_i$ is the embedding learned by the DETR encoder at the $i-th$ position of the prompt $v^*$, and $W_1, W_2, W_3$ are learned weight matrices, and $R$ is the ReLU activation function. 

\begin{equation}
\label{eqn:catr-ce}
\mathcal{L}_{1} = \min \sum_{i=1}^{l}-\mathbb{I}\left(t_{i}\right) \log p\left(t_{i}=t_{i}^{*}\right)
\end{equation}

We train the model by minimizing the sum of the negative cross-entropy over all positions, as depicted in Eq. \ref{eqn:catr-ce}. Here, $t_i^*$ is the probability distribution from Eq. \ref{eqn:catr-output}, and $t_i$ is a one-hot vector encoding the correct token appearing in a caption, and $\mathbb{I}$ is an indicator function that is 1 when $t_i = \text{\texttt{[PAD]}}$ and 0 otherwise. 

\subsection{Fusion Through The Auxillary Sentence}
\label{sec:aux-sentence-fusion}
Once the caption transformer described in Section \ref{par:catr} has been trained, we can use it to translate input images into natural language descriptions of the image. We adapt the auxillary question \cite{sunUtilizingBERTAspectBased2019, devlinBERTPretrainingDeep2019a} mechanism for this. In the auxillary sentence method, BERT is used in sentence-pair classification mode. In sentence-pair classification mode, input to BERT takes the sentence-pair form
\begin{table*}[]
\caption{Dataset Stats}
\label{tab:main-ds-stats}
\begin{tabular}{@{}lllllllllllll@{}}
\toprule
& \multicolumn{6}{c}{\texttt{TWITTER-15}}  & \multicolumn{6}{c}{\texttt{TWITTER-17}}  \\ 
 
\cmidrule(l{0.5em}r{0.5em}){2-7}  
\cmidrule(l{0.5em}r{0.5em}){8-13} 
Split           & Negative   & Neutral & Positive & Total & \# Targets & Length & Negative   & Neutral & Positive & Total & \# Targets & Length \\
 \midrule
Train      & 368        & 1883    & 928      & 3179  & 1.3             & 16.7   & 416        & 1638    & 1508     & 3562  & 1.4             & 16.2   \\
Validation & 149        & 679     & 303      & 1122  & 1.3             & 16.7   & 144        & 517     & 515      & 1176  & 1.4             & 16.4   \\
Test       & 113        & 607     & 317      & 1037  & 1.3             & 17     & 168        & 573     & 493      & 1234  & 1.4             & 16.4   \\ \bottomrule
\end{tabular}
\end{table*}
\begin{table}[]
\caption{Bloomberg Dataset Stats}
\label{tab:bloomberg-ds-stats}
\begin{tabular}{@{}lcccc@{}}
\toprule
  Split         & Text Represented & Text Unrepresented & Total & Length \\
           \midrule
Train      & 927                       & 1324                          & 2251  & 11.7   \\
Validation & 232                       & 332                           & 564   & 11.7   \\
Test       & 290                       & 414                           & 704   & 11.4  \\
\bottomrule
\end{tabular}
\end{table}
\begin{equation}
\label{eqn:sentence-pair}
  \texttt{[CLS]} t^A_1, t^A_2 \ldots t^A_{\operatorname{A.len}} \texttt{[SEP]} t^B_1, t^B_2 \ldots t^B_{\operatorname{B.len}} \texttt{[PAD]} \ldots \texttt{[PAD]}  
\end{equation}
where $t^A_i$ are the tokens of sentence A, and $t^B_i$ are the tokens of sentence B. For aspect sentiment classification, Sentence A is typically the text to be classified, and Sentence B contains auxiliary information used to prime BERT, such as the name of the desired sentiment target. Instead of encoding label information in the auxillary question as in BERT-Pair \cite{sunUtilizingBERTAspectBased2019}, we enrich it with multimodal information. 
Specifically, we concatenate the tokens of the sentiment target with the tokens of the image description predicted by the caption transformer, thus creating a multimodality-enriched auxillary sentence.
We then arrange the the text content and the auxillary sentence into sentence-pair classification mode (Eq. \ref{eqn:sentence-pair}) with the auxillary sentence as Sentence B. We forward the resulting sequence through the BERT encoder, and use the pooler output of the \texttt{[CLS]} token, following previous work \cite{liExploitingBERTEndtoEnd2019, yuAdaptingBERTTargetOriented2019, sunUtilizingBERTAspectBased2019}. Let the $H^\texttt{[CLS]} \in \mathbb{R}^{768}$ be the pooler output. We can then compute the probability $y \in \mathbb{R^3}$ of negative, neutral, or positive sentiment towards the target as follows:

\begin{equation}
\label{eqn:bert-output}
   p(y \mid H^\texttt{[CLS]})=\operatorname{softmax}\left(\theta_{Linear} \operatorname{Dropout}(H^\texttt{[CLS]})\right) 
\end{equation}

Here $\theta_{Linear} \in \mathbb{R}^{3\times768}$, and is learned by backpropagation. We learn $\theta_{Linear}$ by fine-tuning the BERT encoder alongside Eq. \ref{eqn:bert-output} using the standard cross-entropy loss.
\begin{equation}
\mathcal{L}_2=-\frac{1}{|D|} \sum_{j=1}^{|D|} \log p\left(y^{j} \mid H^{\texttt{[CLS]}j}\right)
\end{equation}

\section{Experiments}
We carry out experiments to answer the following research questions:
\begin{itemize}
    \item \textbf{RQ1}: Does our approach using input-space translation improve over a purely text-based language model, and is it competitive with other multimodal and unimodal approaches?
    \item \textbf{RQ2}: Does the image description help BERT to understand the relationship of the image to the tweet, and the target to the image? 
    \item \textbf{RQ3}: How does the amount of multimodal information affect accuracy of predictions in our model?
    \item \textbf{RQ4}: What are the failure modes of powerful state of the art unimodal models on short-text social media data?
\end{itemize}
\subsection{Experiment Settings}
To evaluate our research questions, we use two benchmark datasets for target oriented multimodal sentiment classification, Twitter-15 and Twitter-17, both introduced by \cite{yuAdaptingBERTTargetOriented2019} and building on earlier work \cite{zhangAdaptiveCoattentionNetwork, luVisualAttentionModel2018}. 
Twitter-15 and Twitter-17 consists of multimodal tweets, where each multimodal tweet consists of text, an image posted alongside the tweet, targets within the tweet, and the sentiment of each target. 
Each target is given a label from the set $\{\texttt{negative}, \texttt{neutral}, \texttt{positive}\}$, and the task is a standard multi-class classification problem. 
The third dataset we use is the Bloomberg Twitter-Image Relationship dataset \cite{vempalaCategorizingInferringRelationship}. This dataset also consists of multimodal tweets, but the task for this dataset is a binary classification task, where the tweet-image pair must be classified into two classes: \texttt{TEXT IS REPRESENTED}, indicating that one or more entities in the tweet appear in the image, and \texttt{TEXT NOT REPRESENTED}, indicating that no entities from the tweet appear in the image. To train the caption transformer, we use the MS-COCO \cite{linMicrosoftCOCOCommon2014} dataset, but as our goal is not image captioning, we do not conduct an experimental evaluation for image captioning. \textbf{All hyperpameters are described in Table \ref{tab:hyperparams}}. Most hyperparameters are directly borrowed from DETR \cite{carionEndtoEndObjectDetection2020, uppalSaahiluppalCatr2021} or TomBERT \cite{yuAdaptingBERTTargetOriented2019}. \textbf{Our method introduces no new hyperparameters}. \textbf{We report the average of 5 independent training runs for all our models, and report standard deviations whereever we claim SOTA.}

\subsection{ Multimodal Target Oriented Sentiment Analysis Performance (RQ1)}
We compare three configurations of our approach: (1) \textit{LF-CapTrBERT}, which performs late fusion by skipping the construction of the auxillary sentence and encoding the caption and tweet separately. (2) \textit{EF-CapTrBERT}, which is the model described in Fig. \ref{fig:main-arch-diag} and Section \ref{sec:aux-sentence-fusion}. (3) \textit{EF-CapTrBERT-DE}, where we initialize the BERT layers with domain-specific weights from \cite{nguyenBERTweetPretrainedLanguage2020} but leave the caption transformer unchanged to test if the model can overcome domain mismatch. For text only models, we compare to the baseline BERT model, as well as BERT+BL, which is BERT with another BERT layer stacked on it, and \textit{MGAN} \cite{fanMultigrainedAttentionNetwork2018}, which uses a multi-grain attention network for aspect understanding. We also compare with \textit{BERT-Pair-QA}\cite{sunUtilizingBERTAspectBased2019}, which uses the auxillary question method to obtain SOTA on SemEval 2014 Task 4. For multimodal comparisons, we compare to \textit{TomBERT} \cite{yuAdaptingBERTTargetOriented2019}, which is the current state of the art for target-oriented multimodal sentiment classification, as well as \textit{mPBERT} \cite{yuAdaptingBERTTargetOriented2019}. We also compare with a strong baseline: \textit{Res-BERT+BL}, which directly applies crossmodal attention to ResNet input features and the language features without any extra modifications. 

\begin{table}[]
\caption{Multimodal Aspect Sentiment Classification. $\pm$ are standard deviations across five runs.}
\label{tab:main-ds-perf}
\begin{tabular}{@{}lllll@{}} 
         & \multicolumn{2}{c}{\texttt{TWITTER-15}}  & \multicolumn{2}{c}{\texttt{TWITTER-17}}          \\
\cmidrule(l{0.5em}r{0.5em}){2-3}  
\cmidrule(l{0.5em}r{0.5em}){4-5} 
 Model        & Accuracy   & Mac-F1 & Accuracy   & Mac-F1 \\
 \cmidrule{2-5} \cmidrule{1-1}
 & \multicolumn{4}{c}{Image Only} \\
 \cmidrule{2-5}
RES-Target      & 59.88      & 46.48    & 58.59      & 53.98    \\
\midrule
 & \multicolumn{4}{c}{Text Only} \\
  \cmidrule{2-5}
 MGAN            & 71.17      & 64.21    & 64.75      & 61.46    \\
BERT            & 74.25      & 70.04    & 68.88      & 66.12    \\
 BERT+BL         & 74.25      & 70.04    & 68.88      & 66.12    \\
 BERT-Pair-QA    & 74.35      & 67.7     & 63.12        & 59.66      \\
%  \cmidrule{2-5}
\midrule
 & \multicolumn{4}{c}{Text and Images} \\
 \cmidrule{2-5}
Res-MGAN        & 71.65      & 63.88    & 66.37      & 63.04    \\
 Res-BERT+BL     & 75.02      & 69.21    & 69.2       & 66.48    \\
 mPBERT (CLS)    & 75.79      & 71.07    & 68.8       & 67.06    \\
 TomBERT (FIRST) & 77.15      & 71.75    & 70.34      & 68.03   \\
%   \cmidrule{2-5}
\midrule
 & \multicolumn{4}{c}{Proposed Model Configurations} \\
 \cmidrule{2-5} 
% \midrule
 LF-CapTrBERT    & 76.89      & 72.14    & 68.83      & 66.54    \\
 
 EF-CapTrBERT    & \textcolor{red}{\textbf{78.01}}     & \textcolor{red}{\textbf{73.25 }}   & 69.77        &    68.42   \\
                 & $\pm$ 0.34       & $\pm$ 0.36     & $\pm$  0.16      & $\pm$ 0.48    \\
 EF-CapTrBERT-DE & \textcolor{red}{\textbf{77.92}} & \textcolor{red}{\textbf{73.9}} & \textcolor{red}{\textbf{72.3}} & \textcolor{red}{{\textbf{70.2}}} \\
                 & $\pm$ 0.83 & $\pm$ 0.82 & $\pm$ 0.27 & $\pm$ 0.15 \\
                 \bottomrule
\end{tabular}
\end{table}

\subsubsection{Results \& Discussion} The results are reported in Table \ref{tab:main-ds-perf}. Two configurations of our model, EF-CapTrBERT and EF-CapTrBERT-DE, outperform all text-only models by a substantial margin. Against multimodal models, both EF-configurations outperform the field. However, TomBERT is quite competitive, and only EF-CapTrBERT-DE substantially outperforms TomBERT. \textbf{The comparison between LF-CapTrBERT and the EF-configurations is an ablation.} 
LF-CapTrBERT, does not use the auxillary sentence mechanism and instead encodes the tweet and caption separately, followed by a concatenation, performs much worse than the EF-configurations which use the auxillary sentence. Finally, we note that the stability of EF-CapTrBERT-DE is lower than that of EF-CapTrBERT, which we conjecture is due to the domain mismatch between the caption transformer and the language model in EF-CapTrBERT-DE. 
\begin{table}
\caption{The proposed model's performance improvement over the text-only approach on the Bloomberg Twitter-Image relationship dataset. Standard deviations across 5 runs are reported.}
\label{tab:bloomberg-perf}
\begin{tabular}{@{}lllc@{}}
\toprule
               & Accuracy & Weighted-F1   & Modalities \\ \midrule
Vempala '19 \cite{vempalaCategorizingInferringRelationship}       &    -      & 0.58 & Text+Image  \\

BERT \cite{devlinBERTPretrainingDeep2019a}          & 0.611    & 0.602 & Text\\
 & $\pm$ 0.021    & $\pm$ 0.014 \\
Proposed Method     & \textbf{0.648} (\textcolor{ForestGreen}{+6\%})    & \textbf{0.640} (\textcolor{ForestGreen}{+6.3\%}) & Text + Image\\
 & $\pm$ 0.011    & $\pm$ 0.013 \\
\bottomrule
\end{tabular}
\end{table}
\subsection{Image-Relationship Understanding (RQ2)}
In this experiment, we use the Bloomberg Twitter Image Relationship Dataset (Table \ref{tab:bloomberg-ds-stats}) to answer whether the language model is able to learn a correspondence between the image description, the target, and the tweet. We compare against text-only BERT, and use EF-CapTrBERT as our experimental configuration. We report accuracy and weighted-F1 for consistency with \cite{vempalaCategorizingInferringRelationship}.
\subsubsection{Results \& Discussion} The results can be seen in Table \ref{tab:bloomberg-perf}. The text-only BERT model only outperforms the memory-based model in \cite{vempalaCategorizingInferringRelationship} by a small amount. However, adding multimodal image context with EF-CapTrBERT results in a substantial performance improvement. This shows that EF-CapTrBERT is able to fuse image and text well enough to improve over a purely-text model to understand when it needs to seek additional context from the image.

\subsection{Further Analysis RQ3 + RQ4}
\begin{figure}
    \centering
    \includegraphics[width=\linewidth]{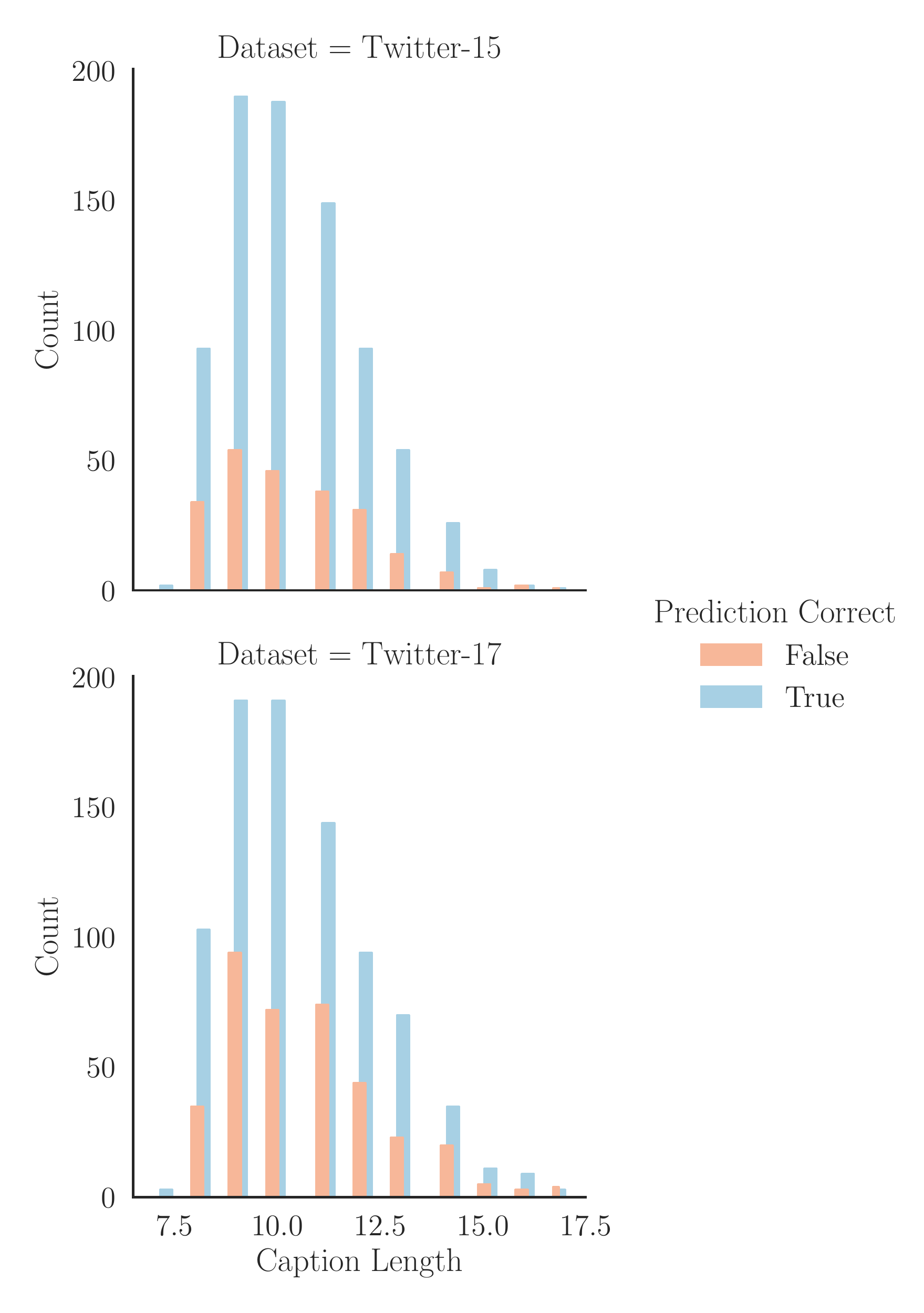}
    \caption{Effect of description length on accuracy.}
    \label{fig:cap-len-acc}
\end{figure}

\begin{figure*}
    \centering
    \includegraphics[width=\textwidth]{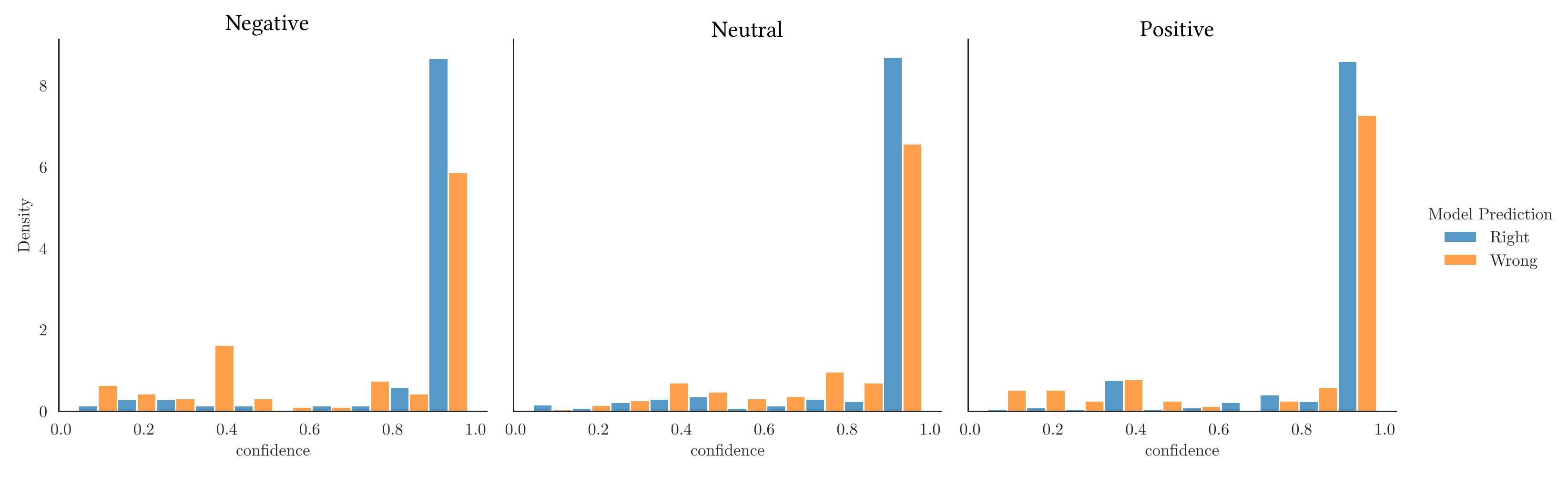}
    \caption{We investigate why state of the art aspect sentiment analysis models, specifically BERT-Pair-QA, experience degraded performance in the social media setting, and find that their confidences are badly calibrated in the social media setting.}
    \label{fig:qc-bert-exposed}
\end{figure*}

\subsubsection{\textbf{RQ3: Effect of image description length on prediction accuracy?}}
We compute the length of each caption, which can be considered a weak proxy for the amount of information in a caption. Then, we bin the captions by length and count the number of correct and incorrect predictions for each bin. The results can be seen in Fig \ref{fig:cap-len-acc}. Surprisingly, we find that there is no strong relationship between the descriptiveness of the image caption and performance.
Intuitively, we might expect longer captions to be more informative, and lead to higher accuracy. 
However, in practice we find that the longer captions also include more noise and more spurious objects, thus counteracting their potential for higher information.
\subsubsection{\textbf{RQ4: Why does BERT-Pair-QA's performance degrade badly on tweets?}}
BERT-Pair-QA is a state of the art and highly effective method for unimodal aspect sentiment analysis.
Surprisingly, it performs badly on Twitter data, even though it exhibits strong performance in other domains. 
BERT-Pair-QA utilizes an auxiliary sentence like our EF-CaTrBERT and EF-CaTrBERT-DE models. 
However, unlike our method, it converts the 3-class sentiment classification problem into a 3-way binary classification problem and uses the confidence levels of each binary classifier to make a choice.
We examine the confidence levels of a BERT-Pair-QA model, specifically 
$$
   \operatorname{Pool}_{OUT} =\operatorname{softmax}\left(\mathbf{W}^{\top} \operatorname{Dropout}(H^\texttt{[CLS]})\right) 
$$
when $H^\texttt{[CLS]}$ contains different label queries in the auxillary sentence. 
We find that the model consistently overestimates probabilities, and this is the reason for performance degradation. The results of our analysis can be seen in Fig \ref{fig:qc-bert-exposed}. One explanation for why this happens is because BERT-Pair-QA was developed for review texts, which tend to be longer than tweets and more informative, so confidence levels may be more trustworthy for reviews than tweets. 
\begin{table}[]
  \caption{Hyperparameters}
  \label{tab:hyperparams}
    \begin{tabular}{@{}lcc@{}}
    \toprule
    Hyperparameter              & Value                & Source         \\ \midrule
    % \multicolumn{3}{c}{Language   Transformer}                         \\
    % \midrule 
    Learning Rate               & 5e-05                & \cite{liExploitingBERTEndtoEnd2019} \\ 
    Attention Heads             & 12                   & \cite{devlinBERTPretrainingDeep2019a} \\ 
    Pooler Dropout              & 0.1                  & Grid Search               \\ 
    Batch Size/Epochs                 & 16/6                   & Grid Search               \\ 
    % Epochs                      & 6                    & Grid Search               \\ 
    Tokenizer Max Length            & 80                   & \cite{liExploitingBERTEndtoEnd2019}  \\ 
    Encoder/Decoder Layers      & 12                   & \cite{devlinBERTPretrainingDeep2019a}  \\ 
    Optimizer/Scheduler                   & AdamW/Linear                & \cite{devlinBERTPretrainingDeep2019a} \\ 
    % Scheduler                   & Linear               & \cite{devlinBERTPretrainingDeep2019a}  \\ 
    Caption Transformer & CATR  & \cite{uppalSaahiluppalCatr2021} \\
    Language Encoder            &bert-base-uncased & \cite{devlinBERTPretrainingDeep2019a} 
    % CNN                         & Resnet101            & \cite{heDeepResidualLearning2015a}         
    \\ \bottomrule
    \end{tabular}
\end{table}

\section{Conclusion}
We identify a number of challenges (high variance in usefulness of the visual modality, scene complexity, and short texts) that set the new field of multimodal aspect/target sentiment analysis apart from multimodal sentiment analysis and aspect/target sentiment analysis. 
We develop an approach, EF-CaTrBERT that uses translation in the input space to translate images into text, followed by multimodal fusion using an auxillary sentence input to the encoder of a language model. We justify the auxillary sentence fusion by showing our LF-CaTrBERT configuration, which lacks the auxillary sentence, cannot match the performance of the EF-CaTrBERT configuration. We show that our input translation significantly improves performance on the task of detecting if entities in tweets are represented in images. We achieve state of the art performance on target/aspect sentiment analysis on the Twitter-15/17 datasets without modifying the architecture of our pretrained BERT language model. We investigate and explain the performance degradation of the otherwise powerful BERT-Pair-QA model on tweets.

% Please add the following required packages to your document preamble:
% \usepackage{booktabs}

% Please add the following required packages to your document preamble:
% \usepackage{booktabs}

% Please add the following required packages to your document preamble:
% \usepackage{booktabs}

%%
%% The acknowledgments section is defined using the "acks" environment
%% (and NOT an unnumbered section). This ensures the proper
%% identification of the section in the article metadata, and the
%% consistent spelling of the heading.
\begin{acks}
We are grateful to Daniele Magazzeni and Robert Tillman at JP Morgan AI Research for many suggestions and thought-provoking conversations which helped shape this paper.
This work was supported by a JP Morgan Faculty Research Award.
\end{acks}

%%
%% The next two lines define the bibliography style to be used, and
%% the bibliography file.
\bibliographystyle{ACM-Reference-Format}
\bibliography{sample-base}

%%
%% If your work has an appendix, this is the place to put it.

\end{document}